\documentclass[runningheads]{llncs}

\usepackage{eccv}

\usepackage{eccvabbrv}

\usepackage{graphicx}
\usepackage{booktabs}

\usepackage[accsupp]{axessibility}  

\usepackage[pagebackref,breaklinks,colorlinks,citecolor=eccvblue,linkcolor=abbrcolor,urlcolor=abbrcolor,bookmarks=false]{hyperref}

\usepackage{orcidlink}

\usepackage{graphicx}
\usepackage{amsmath}
\usepackage{amssymb}
\usepackage{booktabs}
\usepackage[dvipsnames]{xcolor}
\usepackage{makecell}
\usepackage{amssymb}
\usepackage{pifont}
\usepackage{diagbox}
\usepackage{multirow}
\usepackage{amsmath}
\usepackage{bbm}
\usepackage{fancyhdr}
\usepackage{color}
\usepackage[accsupp]{axessibility}  
\usepackage[most]{tcolorbox}
\usepackage{ragged2e}
\usepackage{tabularx}
\usepackage[frozencache,cachedir=minted-cache]{minted} 
\usepackage{wrapfig} 
\definecolor{cvprblue}{rgb}{0.21,0.49,0.74}
\definecolor{abbrcolor}{HTML}{990000}
\definecolor{linkcolor}{rgb}{0.79,0.51,0.26}
\definecolor{jinqilightgray}{HTML}{F5F5F5}
\definecolor{jinqilightorange}{HTML}{FFE7CC}
\definecolor{jinqilightblue}{HTML}{DBE9FC}
\definecolor{jinqilightpurple}{HTML}{E1D6E8}

\definecolor{jinqidarkgray}{HTML}{666666}
\definecolor{jinqidarkorange}{HTML}{D79C00}
\definecolor{jinqidarkblue}{HTML}{6C8EBF}
\definecolor{jinqidarkpurple}{HTML}{9674A6}
\definecolor{jinqidarkgreen}{HTML}{70AD47}
\definecolor{jinqidarkred}{HTML}{990000}

\newcommand{\ourframework}{CKPT\xspace}
\DeclareMathOperator*{\argmax}{argmax}
\DeclareMathOperator*{\argmin}{argmin}

\usepackage{soul}

\newcommand{\myparagraph}[1]{\smallskip\noindent\textbf{#1.}}

\def\abbrstrengthen#1{{\textbf{\textcolor{jinqidarkpurple}{#1}}}}

\begin{document}

\title{Contextual Knowledge Pursuit\\for Faithful Visual Synthesis} 

\titlerunning{Abbreviated paper title}

\author{Jinqi Luo \and
Kwan Ho Ryan Chan \and
Dimitris Dimos \and
René Vidal}

\authorrunning{Luo et al.}

\institute{Center for Innovation in Data Engineering and Science, University of Pennsylvania
\email{\{jinqiluo, ryanckh, dimos, vidalr\}@upenn.edu}}

\maketitle

\begin{abstract}
Modern text-to-vision generative models often hallucinate when the prompt describing the scene to be generated is underspecified. In large language models (LLMs), a prevalent strategy to reduce hallucinations is to retrieve factual knowledge from an external database. While such retrieval augmentation strategies have great potential to enhance text-to-vision generators, existing static top-K retrieval methods explore the knowledge pool once, missing the broader context necessary for high-quality generation. Furthermore, LLMs internally possess rich world knowledge learned during large-scale training (parametric knowledge) that could mitigate the need for external data retrieval. This paper proposes \abbrstrengthen{C}ontextual \abbrstrengthen{K}nowledge \abbrstrengthen{P}ursui\abbrstrengthen{T} (\ourframework), a framework that leverages the complementary strengths of external and parametric knowledge to help generators produce reliable visual content. Instead of the one-time retrieval of facts from an external database to improve a given prompt, \ourframework uses (1) an LLM to decide whether to seek external knowledge or to self-elicit descriptions from LLM parametric knowledge, (2) a knowledge pursuit process to contextually seek and sequentially gather most relevant facts,
(3) a knowledge aggregator for prompt enhancement with the gathered fact context, and
(4) a filtered fine-tuning objective to improve visual synthesis with richer prompts.
We evaluate \ourframework across multiple text-driven generative tasks (image, 3D rendering, and video) on datasets of rare objects and daily scenarios. Our results show that \ourframework is capable of generating faithful and semantically rich content across diverse visual domains, offering a promising data source for zero-shot synthesis and filtered fine-tuning of text-to-vision generative models.

\end{abstract}

\section{Introduction}
\label{sec:introduction}

Recent years have witnessed remarkable advancements in text-to-vision generative models that take textual prompts as inputs to generate various visual outputs (e.g., images \cite{rombach2021highresolution, podell2023sdxl}, 3D rendering \cite{poole2022dreamfusion,wang2023prolificdreamer}). Successful generation from such models hinges on their ability to precisely align and map textual input to corresponding visual representations. With an object in mind, say a dog, an example of a straightforward prompt is ``Generate an image of a dog.'' Nonetheless, when the object is out-of-domain or slightly more complex, short and simple prompts pose challenges, causing issues such as hallucinations and erroneous generations.

\begin{wrapfigure}{r}{0.43\textwidth} 
\centering
\includegraphics[width=\linewidth]{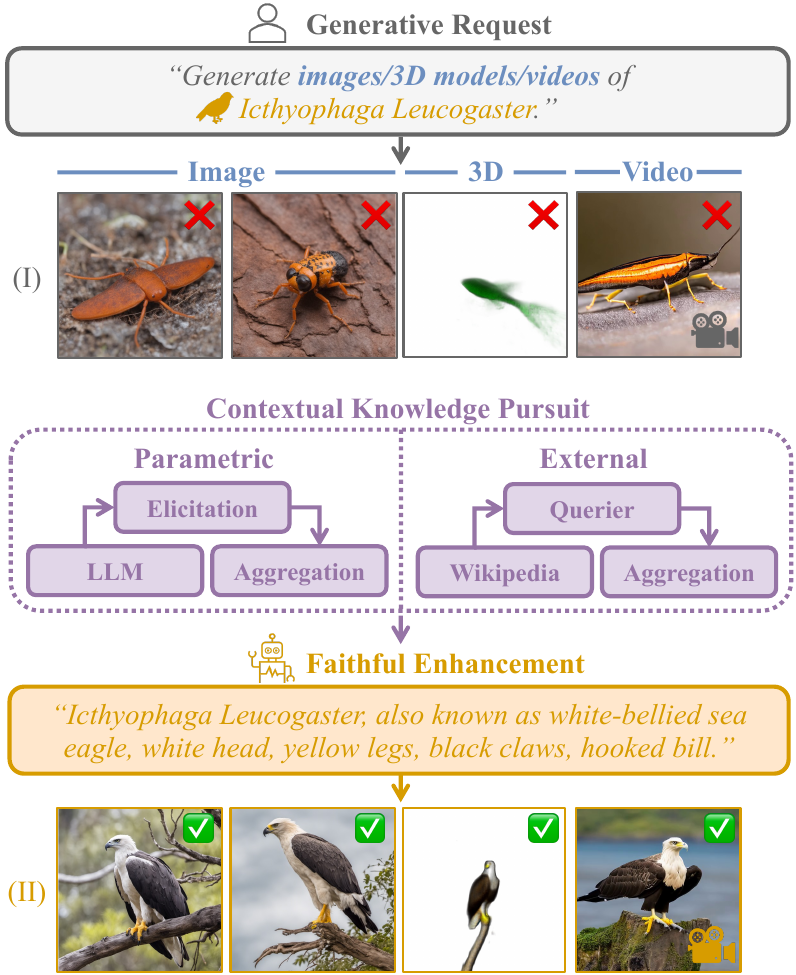}
  \vspace{-5mm}
\caption{Text-driven generative models often produce \textcolor{jinqidarkgray}{(I) unsatisfactory synthesis}. Our proposed framework recursively queries facts in an agent-selected knowledge paradigm to achieve \textcolor{jinqidarkorange}{(II) faithful multimodal synthesis}. 
}
\vspace{-6mm}
\label{fig:teaser_figure}
\end{wrapfigure}
We argue that one reason for poor syntheses of objects such as rare animals, technical or scientific nomenclatures, and ambiguous expressions, is the lack of sufficient detail and information in the text prompt. Namely, the prompt is \textit{under-specified}. 
To address such under-specification during generation, one popular approach is to augment it with a set of relevant knowledge facts. Formally known as Knowledge Retrieval \cite{guu2020realm, lewis2020retrievalaugmented}, a retriever seeks informative references and evidence from a large list of curated, comprehensive, and trusted knowledge sources (e.g., Wikipedia). Notably, prior arts majorly adopt a static top-K query strategy by directly selecting a list of facts with the highest cosine similarity between the embedding of the query and the facts in the database~\cite{Shi2023REPLUGRB, blattmann2022semiparametric}. Nonetheless, this approach has three major disadvantages: First, the method has to index and search through a large knowledge base (e.g., Wiki-DPR has around 21M fact chunks), making it computationally expensive. Second, the set retrieved via the one-time top-K process may not be maximally informative, lacking exploration in the knowledge pool. Last but not least, high-cost external retrieval is not always a necessary condition because language models possess rich knowledge of the world stored in their parameters (i.e., parametric knowledge), which can be also used to address prompt under-specification. Eliciting descriptive statements from such parametric knowledge is efficient but the correctness is not fully guaranteed. Therefore, an effective knowledge-driven framework should be adaptive to the complexity of the task, using parametric knowledge on easy cases that the language model is most familiar with and utilizing external retrieval for the hard cases of the unfamiliar prompts.

In this work, we propose Contextual Knowledge Pursuit (\ourframework), a dynamic knowledge augmented framework designed to improve the input prompt in an adaptive, recursive, and contextual fashion. \ourframework draws inspiration from Information Pursuit (IP)~\cite{chattopadhyay2023variational, chattopadhyay2022interpretable}, a paradigm that sequentially selects informative features for downstream visual tasks. 
As illustrated in Figure~\ref{fig:teaser_figure}, for each incoming prompt, \ourframework first instructs a language agent to determine its familiarity with the content.
If the prompt is familiar (e.g., `kitchen with stove'), \ourframework performs \textit{Parametric Elicitation}, where the language agent recursively self-elicits a chain of descriptive facts, eliminating the indexing process of the external database. If the prompt is unfamiliar (e.g., `Icthyophaga Leucogaster', the scientific name for white-bellied sea eagle), \ourframework conducts \textit{External Retrieval}, 
recursively gathering facts from the external database, which is slower but more faithful. In both cases, \ourframework maintains an adaptive context for the sequential knowledge querying process, subsequently improving the language model's performance for downstream text-to-vision recaptioning. While \ourframework is capable of improving the faithfulness and quality of visual synthesis during inference in a zero-shot manner, recent literature shows that training multimodal models (e.g., CLIP \cite{radford2021clip}, Stable Diffusion \cite{rombach2021highresolution,podell2023sdxl}) on descriptive captions \cite{nguyen2023improvingbyimagecaptioning,openai2023dalle3} is effective in enhancing textual comprehension and model generalization to unseen captions. This motivates us to propose a filtered fine-tuning framework in which we train text-to-vision generators with filtered prompts-image pairs obtained by \ourframework, further enhancing the faithfulness and quality of visual synthesis.

In summary, we propose a novel framework \ourframework that, depending on the language agent's familiarity, contextually gathers the most relevant facts from the parametric knowledge or external database to improve the faithfulness of text-to-vision generative models. \ourframework has the following advantages with respect~to~the~state~of~the~art:
\vspace{-1mm}
\begin{itemize}
 \item \ourframework intelligently decides knowledge paradigm, adapting to various complexities of the tasks by retrieving facts with external knowledge or parametric elicitation.

 \item \ourframework gathers facts in a recursive, dynamic, and contextual way: the queried fact is appended to a knowledge context used for the next pursuit iteration.
 \item \ourframework serves as a high-quality and fidelity data source that can be used for filtered fine-tuning, improving the syntheses for downstream tasks, and reducing hallucinations with more faithful and detailed captions. 
  
\end{itemize}

\vspace{-4mm}
\section{Related Work}
\label{sec:related_work}
\vspace{-2mm}
\subsection{Language-Conditioned Generative Models}
Synthesis from deep generative models, especially variants of StyleGAN \cite{2019stylegan,Karras2020ada} and Diffusion Models \cite{ddpm,rombach2021highresolution,ddim,dhariwal2021diffusion}, can be manipulated in the latent space \cite{stylespace,kwon2023diffusion} using explicit multimodal conditioning (e.g., descriptive text, visual/symbolic template). Earlier efforts adopt CLIP \cite{radford2021clip} to guide image sampling for downstream tasks \cite{2021StyleCLIP,luo2023zeroshot, Wang_2024_CVPR} or optimize the image generator \cite{rombach2021highresolution,2022diffusionclip} with the user text prompt. 
These zero-shot text-to-image generators can serve to supervise both text-to-3D  \cite{poole2022dreamfusion,wang2023prolificdreamer} and text-to-video  \cite{luo2023videofusion,khachatryan2023text2videozero} tasks. More recent works explore the use of language models to improve visual synthesis: \cite{sun2023gpm} takes diffusion models as the visual decoder for LLM-centered multimodal generation, and \cite{zhang2023controllable, lian2023llmgrounded} enhance compositionality of visual synthesis through conditional templates proposed by language models.
However, the faithfulness of LLM-assisted visual synthesis is vaguely addressed in past work.
Our work, instead, pursues a series of factual knowledge as the context to enhance text-driven synthesis, and it generalizes to multiple modalities in a plug-and-play manner.

\vspace{-4mm}
\subsection{Knowledge Retrieval}
Querying informative textual evidence (e.g., background knowledge, semantic context) from external knowledge bases is widely adopted for faithful text synthesis \cite{guu2020realm,lewis2020retrievalaugmented,Jiang2023ActiveRA}, symbolic reasoning \cite{Hu2023ChatDBAL}, and medical guidelines \cite{zakka2023almanac}. Retrieving image data has been adopted for training diffusion model \cite{blattmann2022semiparametric} and joint multimodal transformer training \cite{yasunaga2022retrievalaugmented}. \cite{Shi2023REPLUGRB} found that seeking external evidence for a black-box LLM (e.g., GPT \cite{openai2023gpt4}) can improve its reasoning. However, these approaches either focus on the text modality only, request training the base model, or query all evidence from a single static query.  \cite{chattopadhyay2023variational} made the discovery that, by sequentially querying the data for informative visual semantics, models can reach more interpretable and confident predictions. Inspired by such a sequential procedure, we propose to recursively seek facts with two knowledge paradigms for multimodal synthesis, boosting information expansion, contextual awareness, and retrieval efficiency.

\vspace{-4mm}
\section{Method}
\label{sec:method}
\begin{figure*}[!t]
    \centering
    \includegraphics[width=\linewidth]{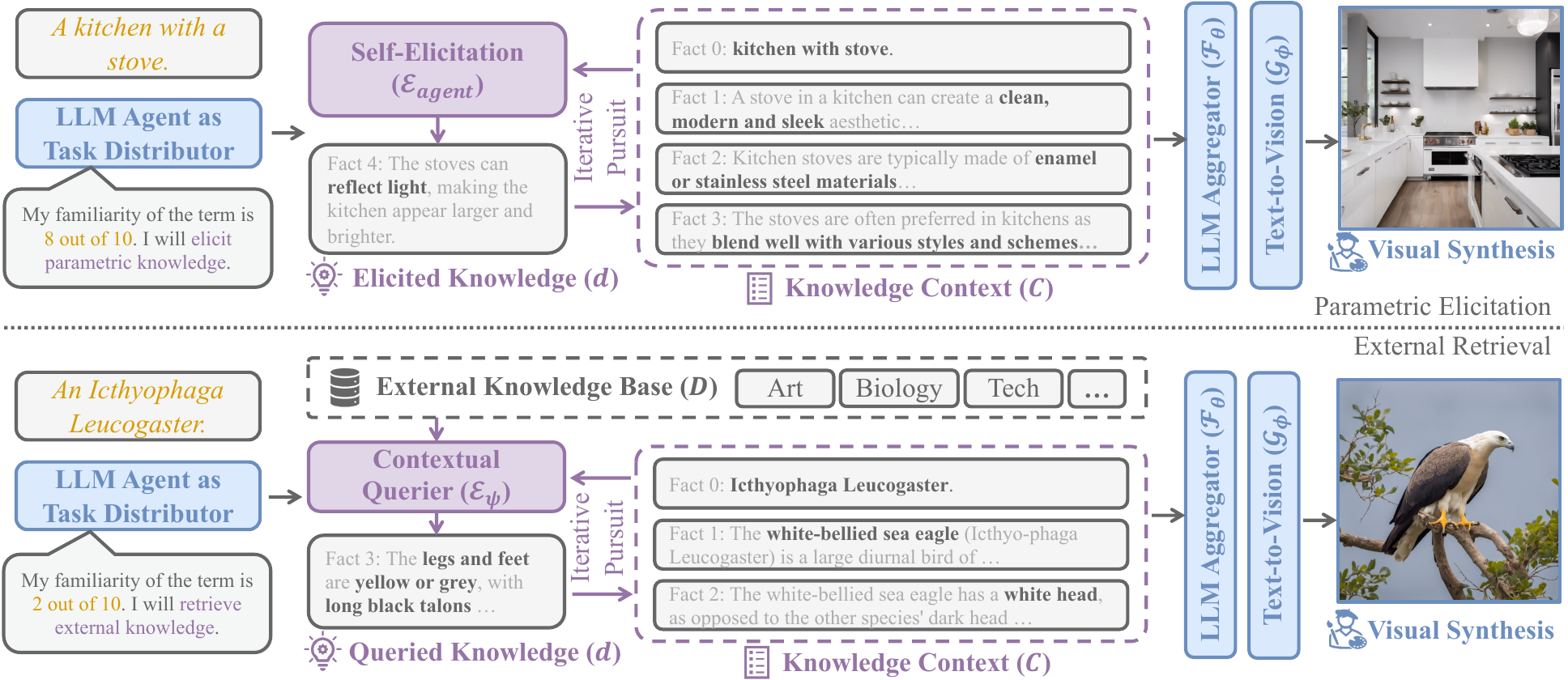}
    \vspace{-6mm}
    \caption{The \ourframework framework. The user inputs a \textcolor{jinqidarkgray}{generic prompt} that lacks details and \ourframework decides the knowledge regime. Then \ourframework recursively picks the most informative given the current state of the \textcolor{jinqidarkpurple}{knowledge context} and appends this fact to update the context. The LLM aggregates the final context to produce a faithfully enhanced caption for \textcolor{jinqidarkblue}{text-driven generators}. }
    \label{fig:method}
    \vspace{-5mm}
\end{figure*}

\vspace{-2mm}
In this section, we present our proposed framework for improving the faithfulness of text-to-vision generative models. In Section~\ref{sec:knowledge_pursuit} we describe our knowledge pursuit method, which iteratively retrieves or elicits a set of facts from an external or parametric source, respectively, depending on the familiarity of a language model with the input prompt. In Section~\ref{sec:language_model_instruction} we describe our knowledge aggregation framework, which uses a language or vision-language model to aggregate the set of gathered facts (knowledge context) into a structured caption. The enhanced caption is then fed to a text-to-vision generator for zero-shot synthesis or filtered fine-tuning (Section~\ref{sec:multimodal_synthesis}). Figure~\ref{fig:method} shows the overview of our framework.

\vspace{-4mm}
\subsection{Knowledge Pursuit}
\label{sec:knowledge_pursuit}
Here we describe our dynamic mechanism to retrieve knowledge from pre-trained language models or external databases in a recursive manner. We assume access to an external knowledge base $D$ that contains a large set of facts, and an instruction-tuned language model $\mathcal{F}_{agent}$. Let $C$ be our knowledge-driven context, a set that will consist of our original task prompt $x$ and a list of facts in the end. Initially, at iteration $k=0$, the knowledge base is the set of all facts, \ie $D_0 = D$, and the context consists of only the task prompt, \ie $C_0 = \{x\}$. 

\myparagraph{Determining Familiarity with Input Prompt}
First, to determine whether CKPT augments knowledge facts using external retrieval or parametric elicitation, we propose a ``get help only when needed'' strategy. Our language model agent will generate an integer score $p_{agent} = \mathcal{F}_{agent}(x)$ to measure its familiarity with the prompt based on its parametric knowledge. 
\vspace{-1mm}
\begin{tcolorbox}[colback=jinqilightblue,colframe=white,halign=justify]
\small
Return an integer from 1 to 10 indicating your level of familiarity with the user input $x$, where 1 means not familiar and 10 means completely familiar.
\end{tcolorbox}
\vspace{-1mm}
Captions with common daily objects (e.g., ‘kitchen with stove’) will receive high scores and will be augmented with knowledge from the language model, while out-of-domain prompts such as wildlife animals and scientific nomenclatures will receive low scores and will be augmented with the external knowledge.

\myparagraph{Contextual Retrieval from External Source} If the agent's familiarity $p_{agent}$ with the input is below a manually-set threshold, it will execute \ourframework with the external database. At a certain step $k$, given the context $C_k$ and the knowledge base $D_k$, the goal is to find a fact that is most relevant for the downstream generative task. We denote it as the fact $d \in D_k$ that maximizes textual relevance between the text embedding of $C_k$ and the text embedding of $d$, \ie:
%
\begin{equation}
    d_{k+1} = \argmax_{d \in D_k} R(\mathcal{E}_{\psi}(d), \mathcal{E}_{\psi}(C_k)), \label{eq:relevance}
\end{equation}
\vspace{-1mm}
where $\mathcal{E}_{\psi}$ is an encoder that maps text tokens to a numeric embedding vector in $\mathbb{R}^p$, $R: \mathbb{R}^p \times \mathbb{R}^p \rightarrow \mathbb{R}^+$ measures the inner-product similarity, and $\mathcal{E}_{\psi}$ to be a pre-trained contrastive text encoder~\cite{izacard2021unsupervised}.
When the total number of words in the context $C$ has not exceeded the maximum input limit of the encoder, we pass the concatenated facts in the context through the encoder as the contextual embedding: $\mathcal{E}_{\psi}(C_k) = \mathcal{E}_{\psi}(\operatorname{concat}\{d_0, \cdots, d_k\})$. Otherwise, we take the average embedding of all individual facts as the representation of the context: $\mathcal{E}_{\psi}(C_k) = \mathbb{E}_{d \in C_k}\left[\mathcal{E}_{\psi}(d)\right]$. Then, the retrieved fact $d_{k+1}$ in the current iteration is \textit{removed} from the knowledge base $D_k$ and \textit{appended to} the current context $C_k$:
\vspace{-1mm}
\begin{align}
        & D_{k+1} = D_k \setminus \{d_{k+1}\}, \\
        & C_{k+1} = C_k \cup \{d_{k+1}\}.
        \vspace{-1.5mm}
\end{align}
\vspace{-5mm}

\myparagraph{Self-Elicitation from Parametric Knowledge} If the agent decides that it is sufficiently familiar to the prompt, \ourframework performs its own parametric knowledge. At each step $k$, given the context $C_k$, we instruct the agent to generate a descriptive fact $d_{k+1} = \mathcal{E}_{agent}(C_k)$ from its implicit knowledge that complements $C_k$ for the downstream generative task. Similar to the external retrieval, after each query, the elicited fact will be appended to the knowledge context to update the context. The parametric paradigm mimics the retrieval paradigm by instructing the agent to read the previously retrieved context and perform self-elicitation to construct a chain of semantic knowledge. 

In both paradigms, the querying process stops after $n$ steps, where $n$ is either a user-defined upper bound of knowledge context size or depends on the maximum number of tokens allowed for the subsequent language model. Alternatively, one can design a stopping criterion, \eg (1) when the textual relevance score (Equation~\ref{eq:relevance}) of any query and the context is below a pre-determined threshold, (2) the language model agent's familiarity score is above a threshold after iterations, or (3) when the embedding of $C_k$ does not change much from one iteration to the next.

As validated in Section~\ref{sec:sequential_knowledge_query}, the contextually recursive retrieval searches for more informative knowledge compared to static top-K retrieval. Using ``Icthyophaga Leucogaster'' as an example, the database's directly related knowledge regarding this string is quite limited. However, ``white-bellied sea eagle'' (which is the bird's common name) is associated with a richer pool of available knowledge. 
Our approach establishes an association between them in the first iteration, and then recursively explores other relevant knowledge based on both strings. The construction of the knowledge context allows for a more expansive and rich exploration of the implicit knowledge distribution. 
\vspace{-4mm}
\subsection{Knowledge Aggregation}
\label{sec:language_model_instruction}
\vspace{-0.5mm}
This section introduces the knowledge aggregation mechanism to enhance the given underspecified captions.
Once we have generated a knowledge-driven context $C$, we instruct an independent LLM $\mathcal{F}_{\theta}$ to further parse and aggregate the context for the downstream text-to-vision generator. 
Our in-context instruction majorly consists of three key components:
\vspace{-1mm}
\begin{tcolorbox}[colback=jinqilightpurple,colframe=white,halign=justify]
\small
Here is a demonstration: \\
Knowledge: $d_1, \dots, d_n$\\
Original Prompt: $x_{\text{demo}}$\\
Enhanced Prompt: $T_{\text{demo}}$
\end{tcolorbox}
\vspace{-4mm}
\begin{tcolorbox}[colback=jinqilightorange,colframe=white]
\small
Read the following knowledge and enhance the user original prompt for the \texttt{<Generator>}. You may add more appearance details, semantic attributes, and fine-grained visual elements.
\end{tcolorbox}
\vspace{-4mm}
\begin{tcolorbox}[colback=jinqilightblue,colframe=white,halign=justify]
\small
If you find some piece of knowledge irrelevant or conflicting to the original prompt, you may ignore the piece. You may also remove meaningless words. You should make the prompt concise, expressive, and accurate.
\end{tcolorbox}
\vspace{-1mm}
\textcolor{jinqidarkpurple}{Parsing Demonstration} guides the language model to digest the curated list of knowledge and enhance the prompt for downstream text-driven generators by providing one in-context example. 
Then, \textcolor{jinqidarkorange}{Enhancement Request} describes the task that the language model should execute. 
We associate the generator information (e.g., \texttt{<Generator>} = ``Stable Diffusion XL'') in the instruction for better generator-aware enhancement.
Last but not least, \textcolor{jinqidarkblue}{Knowledge Rejection} addresses the issue that the queried knowledge is not always helpful to the user's task. This happens if the language agent hallucinates a fact during self-elicitation or the underlying relevance of an external fact is not satisfactory. 
The knowledge rejection rule lets the language model override the knowledge context when it finds conflicts or redundancies.

By concatenating three components (as $I$) with the query knowledge context (as $C$), the language model will generate the enhanced prompt $x^* = \mathcal{F}_{\theta}(I \cup C)$ for the downstream text-driven generative model.

\begin{wrapfigure}{r}{0.5\textwidth} 
\vspace{-8mm}
\centering
\includegraphics[width=\linewidth]{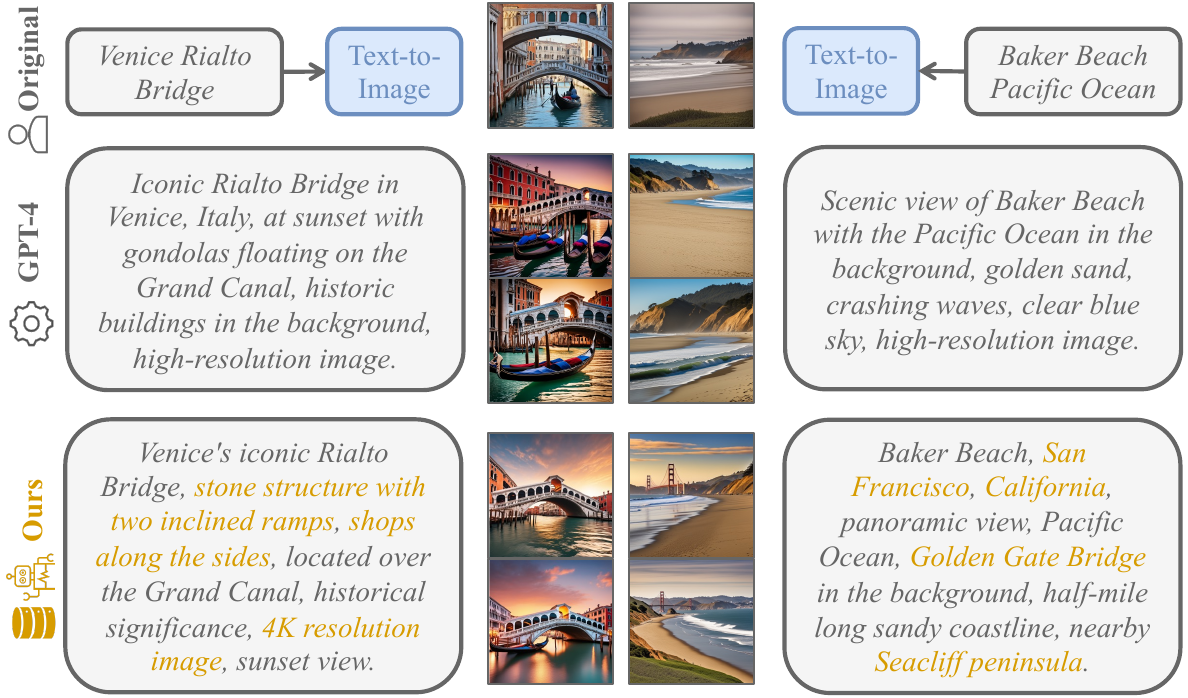}
  \vspace{-5mm}
\caption{Generative captions obtained by prompting GPT-4 and by \ourframework. The former yields generic descriptions, while \ourframework produces a more detailed, precise, and faithful prompt.}
\vspace{-7mm}
\label{fig:prompt_demo}
\end{wrapfigure}

\vspace{-4mm}
\subsection{Zero-Shot Synthesis and Filtered Fine-tuning}
\label{sec:multimodal_synthesis}
\vspace{-1mm}
This section discusses our procedure of zero-shot synthesis and its application in text-to-vision diffusion fine-tuning.
Given enhanced captions $x^*$ by knowledge aggregation, we obtain the zero-shot visual synthesis $y^* = \mathcal{G}_{\phi}(x^*)$, where $\mathcal{G}_{\phi}$ is a pre-trained text-driven generator. Two examples of our enhancement with more faithful details are shown in Figure~\ref{fig:prompt_demo}.

Although zero-shot synthesis of \ourframework already produces high quality visuals, optimizing the visual generator for novel and complex prompts requires further fine-tuning. Recent literature shows that training multimodal models (e.g., CLIP \cite{radford2021clip} or Stable Diffusion \cite{rombach2021highresolution,podell2023sdxl}) on descriptive captions \cite{nguyen2023improvingbyimagecaptioning,openai2023dalle3} is effective in enhancing the textual-visual reasoning capability. 
We pair \ourframework-generated images $y^*$ with enriched prompts $x^*$ to form an enhanced dataset $\mathcal{D^*} = \{(x^*, y^*)\}_{i=1}^N$. We manually curate a subset of $k$ high-quality pairs from $\mathcal{D^*}$ based on visual faithfulness and observered aesthetics. Simultaneously, we rank (enhanced caption, original image) pairs $(\hat{x}, y)$\footnote{To avoid notation confusion, we note that $\hat{x}$ is not necessarily the same set as $x^*$ although they both indicate \ourframework-enhanced captions.} using CLIP scores, removing the $k$ lowest-scoring pairs to refine the original dataset $\mathcal{D}$.
The filtered $\mathcal{D}$ contains real images with corresponding enhanced captions, and the filtered $\mathcal{D^*}$ is a high-quality synthetic dataset of faithful diffusion images generated on descriptive captions.
By fine-tuning text-to-image diffusion models on these two filtered sets jointly, the model's text-vision generative capability will be improved. Given a frozen text-to-image diffusion model $\mathcal{G}_{\phi}$, our training objective of its low-rank adaptor \cite{hu2022lora} $\phi_{lora}$ is:
\begin{align}
\phi_{lora}^* &= \argmin_{\phi_{lora}} \mathbb{E}_{t \sim T} \mathbb{E}_{(\hat{x}, y, x^{*},  y^*) \sim \mathcal{D} \times \mathcal{D^*}}  \notag \\
&\left[ \lVert\epsilon_{\phi + \phi_{lora}}\left(y_t, \hat{x}, t\right)-\epsilon_s (y, t)\rVert^2 + \lVert\epsilon_{\phi + \phi_{lora}}\left(y_t^*, x^*, t\right)-\epsilon_s (y^*, t)\rVert^2\right],
\end{align}
where $\epsilon_{\phi + \phi_{lora}}$ is the score function of $\mathcal{G}_{\phi}$ and $\epsilon_s$ is the noise scheduler.
\section{Experimental Results}
\label{sec:experiment}

We present the empirical validation of the \ourframework enhancement on multiple text-driven generative paradigms. Section~\ref{sec:framework_steup} states the necessary setup of the framework and experiments. Section~\ref{sec:synthesis_visualization} visualizes the multimodal enhancement results, and Section~\ref{sec:effectiveness_evaluation} evaluates the visual synthesis by several benchmarks. Section~\ref{sec:framework_design_analysis} analyzes the design principles and extensibility of our framework.

\vspace{-2mm}
\subsection{Implementation Details}
\label{sec:framework_steup}

\myparagraph{Dataset} We assess the zero-shot effectiveness of our method across image captions from MSCOCO 2017 \cite{lin2014mscoco}, Global Biodiversity Information Facility (GBIF) Taxonomy \cite{gbif2011}, and GUIE LAION-5B \cite{schuhmann2022laion5b,guielaion5b}. For each dataset, we sample 2,000 captions and generate 10 images for each caption. We further test our fine-tuning improvement on the HEIM benchmark \cite{lee2023holistic_heim}, where the evaluation picks 100 prompts per generative aspect. We evaluate the effectiveness of our language querying strategy on sampled questions from MMLU \cite{hendrycks2021mmlu}. The knowledge context size is 8 in \ourframework. The external knowledge base is Wikipedia up to December 2018 with FAISS indexing \cite{johnson2019billion}. 

\myparagraph{Language Model} The knowledge aggregation backbone of \ourframework is the GPT-4 \cite{openai2023gpt4}. Recent literature reports that its behaviors shift over time since the base model behind the interface/API could be changing \cite{chen2023chatgptshift}. To ensure minimal behavior shift over time for better reproducibility, we use the fixed snapshot \texttt{gpt-4-0613} completion API. In Section~\ref{sec:diversity_and_multimodal_generality} (multimodal generality), we embed \ourframework with the multimodal \texttt{gpt-4-vision-preview}.

\myparagraph{Text-Driven Generator} We adopt publicly available generators for different modalities. We use Stable Diffusion (SD) XL \cite{podell2023sdxl} for text-to-image synthesis except for Section~\ref{sec:component_study} (component study) where we start with SD-1.5 \cite{rombach2021highresolution}.  In Section~\ref{sec:diversity_and_multimodal_generality} (multimodal generality), we also experiment with \texttt{dall-e-3} API. The quality evaluation uses the Fréchet Inception Distance (FID) \cite{heusel2017fid} and Inception Score (IS) \cite{salimans2016is}. We resize all images to 224 $\times$ 224 for FID and 299 $\times$ 299 for IS. In the text-to-3D domain, we employ DreamFusion \cite{poole2022dreamfusion} embedded with DeepFloyd IF \cite{DeepFloyd2023} for experiments considering its efficient inference, and ProlificDreamer \cite{wang2023prolificdreamer} for pipeline demonstrations. Video synthesis was performed using ZeroScope \cite{cerspense2023,luo2023videofusion} and Text2Video-Zero \cite{khachatryan2023text2videozero} in the manuscript.

\subsection{Synthesis Visualization}
\label{sec:synthesis_visualization}
This section demonstrates the benefits of knowledge-driven refinement for visual synthesis across multiple modalities.

\begin{figure*}[h]
    \centering
    \includegraphics[width=\linewidth]{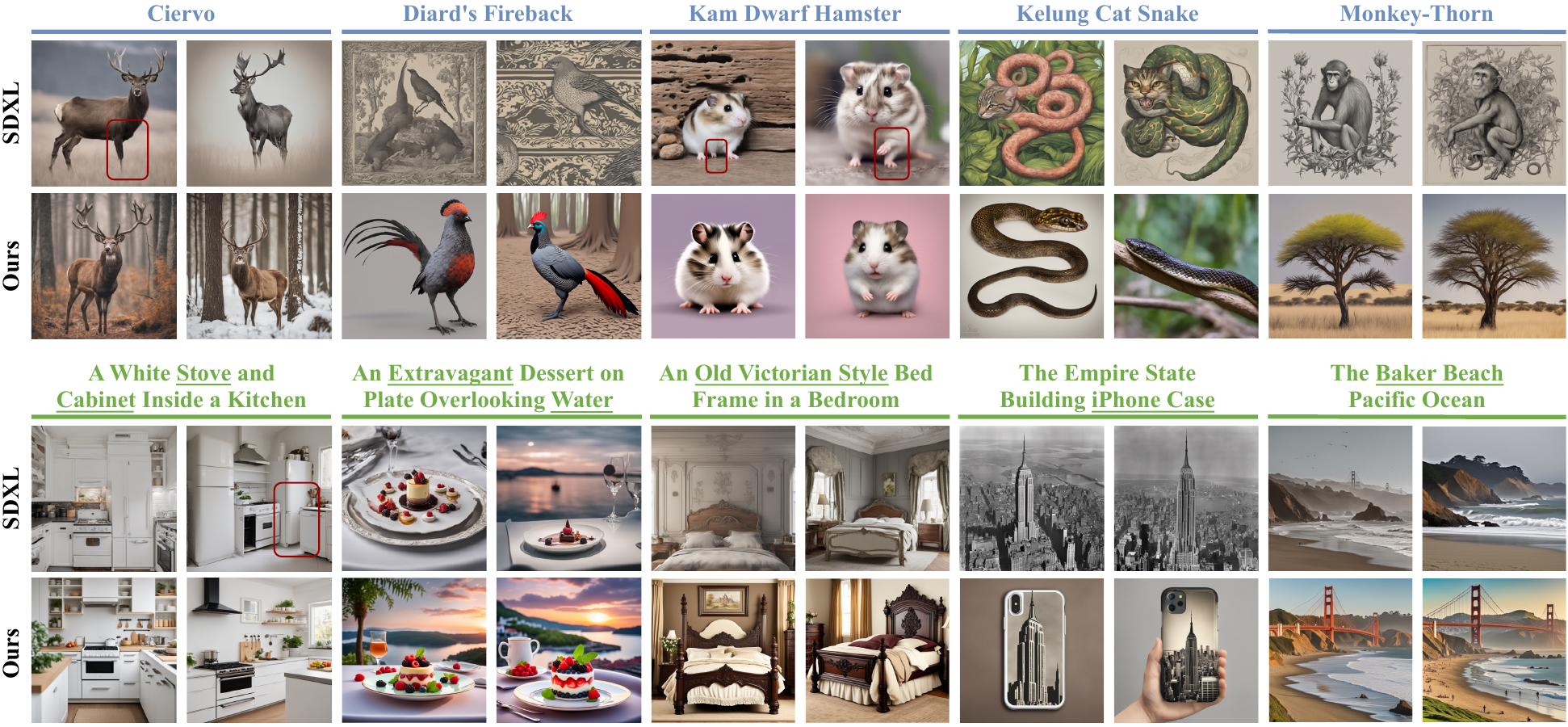}
    \vspace{-5mm}
    \caption{Comparison of images generated from original captions and \ourframework on the GBIF (upper two rows), MSCOCO (three captions on the lower left), and GUIE LAION-5B (two captions on the lower right) datasets. 
    Blue columns demonstrate results from the \textcolor{jinqidarkblue}{external retrieval}, while green columns emphasize the \textcolor{jinqidarkgreen}{parametric elicitation}. We outline \textcolor{jinqidarkred}{deficits in bounding boxes} (e.g., the missing deer legs, the stuck closed fridge door) and \underline{underline notable concepts} that the generator should express sufficiently.}
    \label{fig:text_to_image_all_datasets}
    \vspace{-1mm}
\end{figure*}
\begin{figure*}[!t]
\centering
  \includegraphics[width=\linewidth]{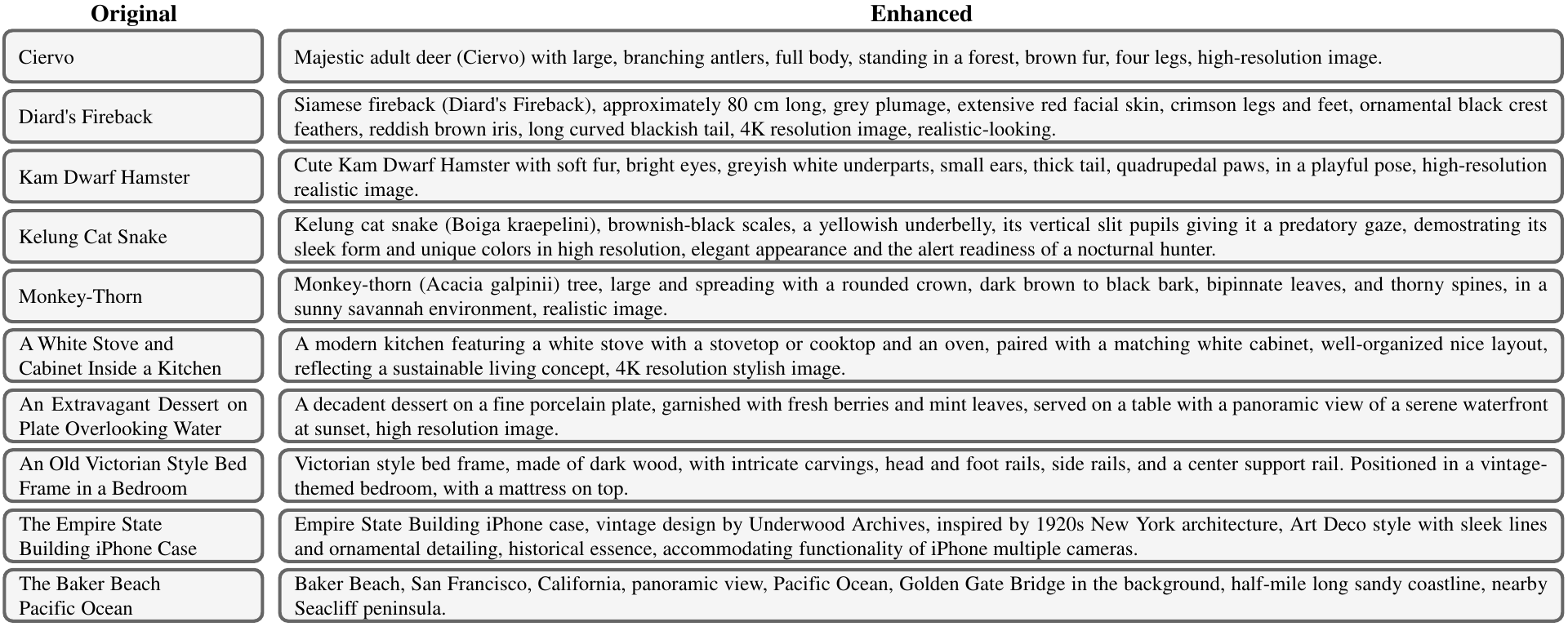}
\vspace{-5mm}
\caption{The \ourframework-enhanced captions of Figure~\ref{fig:text_to_image_all_datasets}. We observe that the descriptive captions generated by our framework well capture core semantics and concepts from contextual knowledge queries. }
\label{fig:text_caption}
\vspace{-2mm}
\end{figure*}

\vspace{-2mm}
\subsubsection{Image Synthesis}

Figure \ref{fig:text_to_image_all_datasets} shows the \ourframework synthesis using captions of creature names (for biological research) from GBIF, scene captions from MSCOCO, and landmark captions from GUIE LAION-5B. We observe that the original stable diffusion XL produces hallucinatory contents. On the other hand, \ourframework improves the synthesis quality (e.g., the legs for deer and hamster), and our knowledge-driven process corrects the ambiguity of the prompts by augmenting with more facts about creatures (e.g., Money-Thorn\footnote{https://www.feedipedia.org/node/352} is a kind of tree from Africa, and Kelung Cat Snake\footnote{https://eol.org/pages/795578} is a species of colubrid snakes). From the MSCOCO captions, we observe that \ourframework produces more fine-grained content such as the reasonable layout of kitchen, the elegance of the decorated dessert, and the Victorian style of the bed. The \ourframework synthesis on LAION-5B captions is more faithful, as the Baker Beach in San Francisco is famous for the joint scenic view with the Golden Gate Bridge, which the SDXL original synthesis often fails to capture. Full \ourframework-enhanced captions are shown in Fig.~\ref{fig:text_caption}. 


\begin{figure*}[!t]
\centering
  \includegraphics[width=\linewidth]{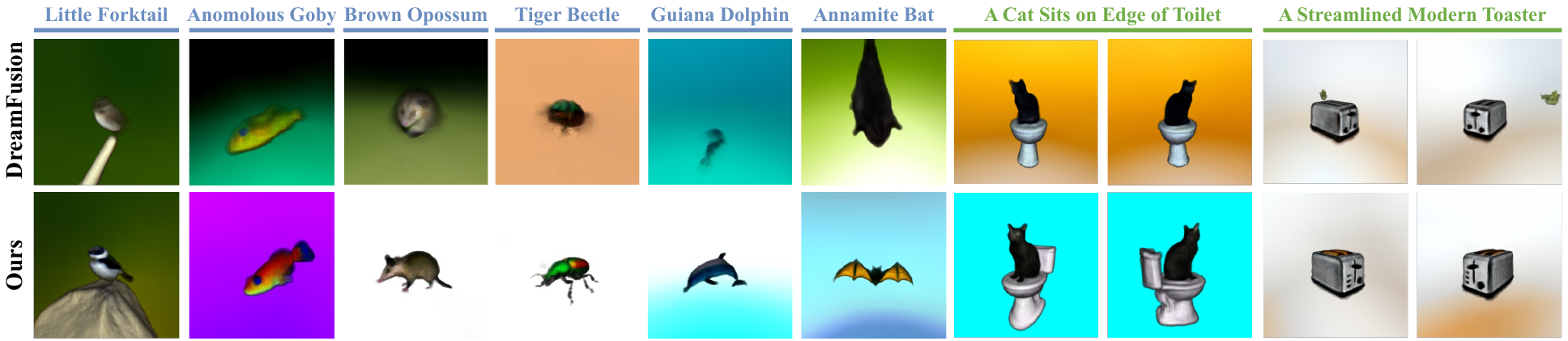}
  \vspace{-5mm}
\caption{Comparison of 3D rendering generated from original captions and \ourframework-external on the \textcolor{jinqidarkblue}{GBIF} and \textcolor{jinqidarkgreen}{MSCOCO} datasets. 
The text-to-3D model is DreamFusion \cite{poole2022dreamfusion} embedded with DeepFloyd-IF \cite{DeepFloyd2023}. 
We observe that our framework enhances the color patterns and semantic contour of rare animal prompts, and improves the synthesis quality with more details on common object prompts.}
\label{fig:text_to_3d}
\vspace{-3mm}
\end{figure*}

\begin{figure*}[!t]
\centering
  \includegraphics[width=\linewidth]{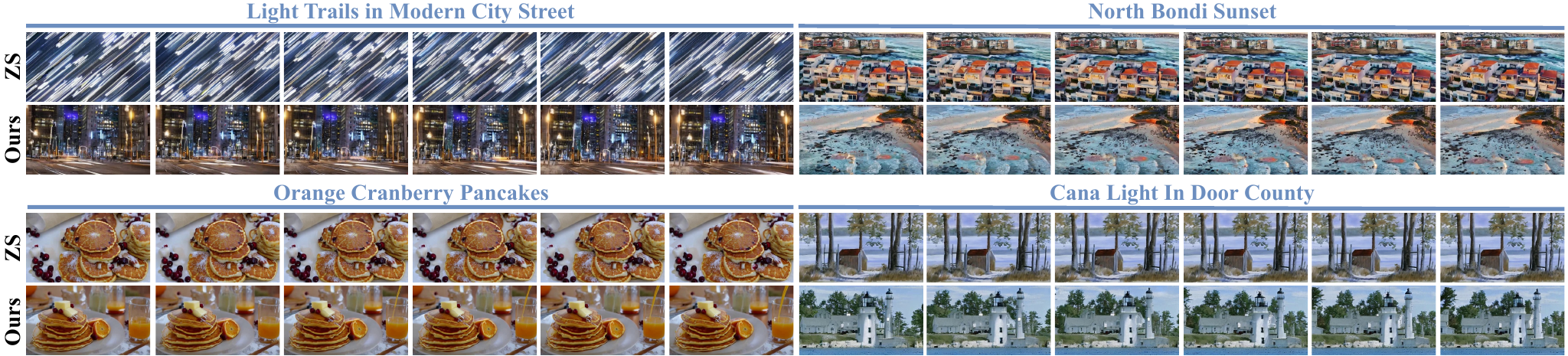}
\vspace{-5mm}
\caption{Comparison of videos generated with original captions and \ourframework-external on the GUIE LAION-5B dataset. The text-to-video model is ZeroScope \cite{cerspense2023}, a watermark-free fine-tuning of VideoFusion \cite{luo2023videofusion}. We can observe that our framework enhances the synthesis quality (e.g., synthesis of street view, correct patterns of pancakes with decomposed oranges), and improves the realism and faithfulness for knowledge-intensive captions (e.g., landmarks like North Bondi Beach in Sydney and Cana Light in Wisconsin).}
\label{fig:text_to_video}
\vspace{-5mm}
\end{figure*}

\subsubsection{3D and Video Synthesis}
\ourframework can enhance textual descriptions to generate fine-grained 3D rendering. Figure \ref{fig:text_to_3d} shows the zero-shot 3D rendering from the GBIF and MSCOCO captions. Specifically, the 3D rendering with generic captions either produces inaccurate depictions or fails to converge. For example, DreamFusion renders a toilet bowl with a missing water tank, and the synthesis of the beetle is ambiguous. \ourframework results, instead, show clear semantic attributes for captions of rare species, and valid appearance with correct shapes on compositional captions.

Figure~\ref{fig:text_to_video} shows the generated video from the LAION captions by ZeroScope. The \ourframework-enhanced videos are more coherent with the captions, and the visual faithfulness is much improved. For example, the Orange Cranberry Pancakes\footnote{https://www.food.com/recipe/cranberry-orange-pancakes-304319} (a breakfast dish with pancake, cranberry, and orange fruit) directly synthesized by ZeroScope is a hallucinatory mixture of an orange and a pancake shape. After the enhancement by \ourframework, the pancakes show adequate shape and a reasonable appearance. The cranberries with butter squares are placed on top of the pancakes, while the orange is placed adjacent to the pancakes. There is also a dynamically pouring orange juice in the video by \ourframework.

\subsection{Quantitative Evaluation}
\label{sec:effectiveness_evaluation}
This section assesses the quality and faithfulness of \ourframework, and benchmark our recursive querying strategy against established baselines.

\begin{table}[!t]
\scriptsize
\centering
  \caption{\label{tab:aesthetic_evaluation} The HEIM \cite{lee2023holistic_heim} aesthetic evaluation on fine-tuned text-to-image generators with training data from different image recaptioners. The aesthetic scores are from the LAION-Aesthetics-V2 predictor. The classifier score is between 1 and 10, higher is better. We can observe that, after fine-tuning on \ourframework synthesis, the generator has the finest aesthetics in all knowledge-intensive subjects. }
\begin{tabular}[width=\linewidth]{@{}lcccc@{}}
\toprule
& \makecell{CUB-200 Bird \cite{WahCUB_200_2011}} & \makecell{Logo \cite{lee2023holistic_heim}} & \makecell{Historical\\Figures \cite{lee2023holistic_heim}} & \makecell{Occupation\\Fairness \cite{Bianchi_2023, Cho2023DallEval}}  \\
\midrule
Vanilla LoRA \cite{hu2022lora} & $5.64 \pm 0.68$ & $3.93 \pm 1.15$ & $5.13 \pm 1.39$ & $5.19 \pm 1.51$\\
InstructBLIP Captioner \cite{dai2023instructblip} & $6.39 \pm 0.73$ & $6.24 \pm 0.73$ & $6.64 \pm 1.04$ & $5.48 \pm 1.15$  \\
LLaVA-1.5 Captioner\cite{liu2023improvedllava} & $6.36 \pm 0.68$ & $5.38 \pm 1.36$  & $6.58 \pm 1.13$ & $5.18 \pm 1.37$ 
 \\
$\displaystyle \text{\ourframework (Ours)}$ & \textbf{7.31 $\pm$ 0.62} & \textbf{7.48 $\pm$  0.70}  & \textbf{7.49 $\pm$  0.96} & \textbf{6.93 $\pm$  0.92}  \\
\bottomrule
\vspace{1mm}
\end{tabular}

\begin{tabular}[width=\linewidth]{@{}lcccc@{}}
\toprule
     & \multicolumn{4}{c}{PartiPrompt \cite{partibenchmark}}  \\
     \cmidrule(lr){2-5}
& \makecell{Arts} & \makecell{Food} & \makecell{Vehicles}& \makecell{World Knowledge} \\
\midrule
Vanilla LoRA \cite{hu2022lora}  & $6.14 \pm 0.94$ & $5.78 \pm 0.84$ & $5.76 \pm 0.96$ & $5.77 \pm 1.15$\\
InstructBLIP Captioner\cite{dai2023instructblip}  & $6.59 \pm 0.93$ & $6.88 \pm 1.03$ & $6.65 \pm 0.92$ & $6.38 \pm 1.07$   \\
LLaVA-1.5 Captioner\cite{liu2023improvedllava}  &  $6.61 \pm 0.83$ & $6.86 \pm 0.94$ & $6.77 \pm 0.89$ & $6.36 \pm 1.24$
 \\
$\displaystyle \text{\ourframework (Ours)}$  & \textbf{7.33 $\pm$  0.82} & \textbf{7.35 $\pm$ 0.95} & \textbf{7.61 $\pm$  0.80} & \textbf{7.57 $\pm$  0.85}  \\
\bottomrule
\end{tabular}
\vspace{-3mm}
\end{table}

\subsubsection{Fine-Tuning}
 We compare the fine-tune of SDXL with data generated by different image recaptioners: (1) LLaVA-1.5-13B \cite{liu2023improvedllava}, (2) InstructBLIP-Vicuna-7B \cite{dai2023instructblip}, (3) \ourframework. All baselines share the same original caption set from the sampled LAION-5B and GBIF datasets. We adopt HEIM \cite{lee2023holistic_heim} to evaluate the aesthetic performance of these fine-tuned generators on eight categories as listed in the Table~\ref{tab:aesthetic_evaluation}. We observe that \ourframework outperforms in all categories across know-intensive, fair, and diverse synthesis.

\subsubsection{Synthesis Faithfulness}

\begin{wrapfigure}{r}{0.48\textwidth} 
    \centering
    \vspace{-8mm}
    \includegraphics[width=\linewidth]{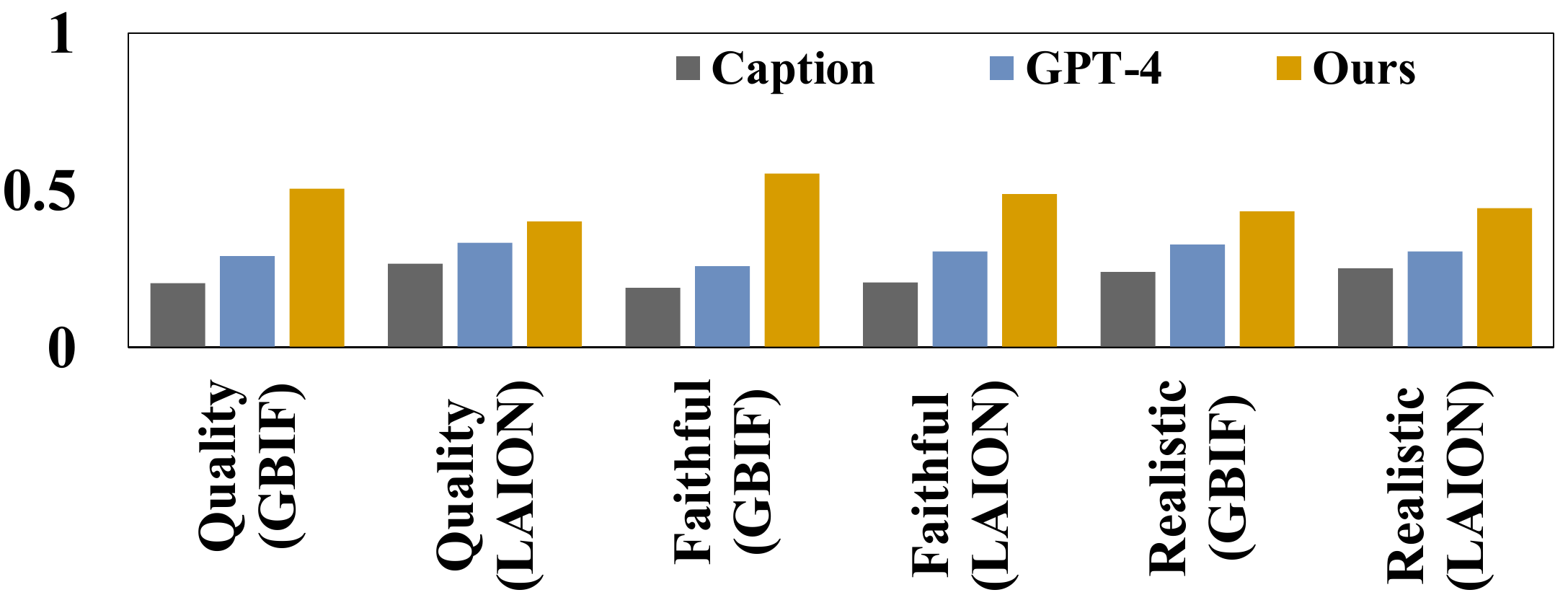}
    \vspace{-5mm}
    \caption{User evaluation on images generated by different approaches. The vertical axis represents the fraction of images
    that are selected as the most high-quality $\setminus$ faithful $\setminus$ realistic. We can observe that our framework outperforms in all three aspects.}
    \vspace{-4mm}
    \label{fig:user_study}
\end{wrapfigure}

We conduct a user study to evaluate the visual soundness of our generative framework. We recruit 10 independent users with Bachelor's degrees to evaluate the faithfulness, quality, and realism (how realistic the image looks) of the images by \ourframework and two baselines. Take the faithfulness evaluation as an example. Each user reads 100 triplets of images by three methods corresponding to the same instance of data. The sequences of the images are randomly shuffled. The user selects the image that is most faithful among all. 
Figure~\ref{fig:user_study} shows the frequency of each method being selected. We can observe that the average rating for \ourframework is higher than the two baselines. This indicates that our knowledge pursuit process effectively enhances the faithfulness and quality of synthesis from the user perspective. 

\begin{table}[!t]
   \centering
      \caption{\label{tab:llm_effectiveness} The question-answer evaluation on different retrieval paradigms. HS indicates the High-School level in MMLU. The numbers are presented as correct answers / total questions. We can observe that \ourframework has the highest correctness in all subjects. }
\begin{tabular}[width=\linewidth]{@{}lccccc@{}}
\toprule
& \makecell{Geography\\(HS)} & \makecell{Global\\Fact} & \makecell{Biology\\(HS)} & \makecell{World\\History} & Total Count ($\uparrow$) \\
\midrule
GPT Closed Book & 21/30 & 9/30 & 20/30 & 23/30 & 73/120 \\
Fact Aggregation & 24/30 & 11/30 & 20/30 & 25/30 & 80/120 \\
REPLUG \cite{Shi2023REPLUGRB} & 23/30 & 13/30 & 22/30 & 24/30 & 82/120 \\
CKPT (Ours) & 24/30 & 13/30 & 23/30 & 25/30 & \textbf{85/120} \\
\bottomrule
\end{tabular}

\end{table}

\vspace{-3mm}
\subsubsection{Effectiveness of the Query Strategy}
\label{sec:sequential_knowledge_query}
This section emphasizes our sequential knowledge pursuit process which dynamically queries each new fact based on the state of knowledge context. 
We compare our query strategy with REPLUG \cite{Shi2023REPLUGRB} and Static Aggregation in the textual question answering task.
REPLUG queries the top-K most informative facts in a static one-time search and makes a prediction by ensembling the LLM posteriors from each fact.
Fact Aggregation, the static query version of \ourframework, samples all top-K facts
in an one-time query or self-elicitation as the context and instructs the LLM to aggregate. 
Table~\ref{tab:llm_effectiveness} shows the results using four subjects from MMLU \cite{hendrycks2021mmlu} that the subject information is relevant to faithful visual synthesis. We sample 30 questions from each subject. We follow REPLUG's setup and choose OpenAI's \texttt{text-davinci-003} legacy API as the language model base for all three methods.
We see that our recursive querying strategy is more performant in knowledge-intensive language tasks, indicating the strong capability to aggregate knowledge for our multimodal synthesis. 
Our observations reveal that when multiple corpora are queried in a single request, the knowledge received tends to overlap, offering little diversity. Instead, our recursive method iteratively introduces new terms into the knowledge context for the next knowledge query.
We hypothesize that the recursive construction of this knowledge context contributes to the superior performance of our paradigm. 

\subsection{Framework Design Analysis}
\label{sec:framework_design_analysis}
This section discusses the modular design of \ourframework, explores variants of foundation model bases, and analyzes how our system is progressively built.

\begin{figure}[!t]
\centering
\begin{minipage}[b]{0.49\linewidth} 
\centering
  \includegraphics[width=\linewidth]{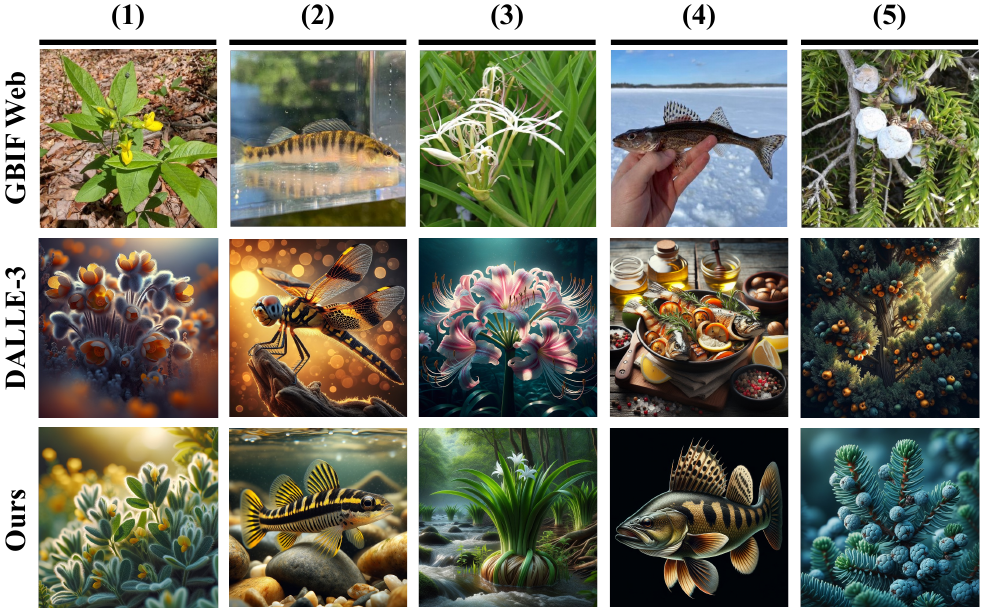}
  \vspace{-2mm}
\caption{The comparisons of real images of the species from the GBIF Taxonomy Website, synthesis by captions, and synthesis by \ourframework-external embedded with \texttt{dall-e-3} API as the generator $\mathcal{G}_{\phi}$. 
}
\label{fig:text_to_image_dalle}
\vspace{-1mm}
\end{minipage}
\hfill
\begin{minipage}[b]{0.49\linewidth}
\centering
  \includegraphics[width=\linewidth]{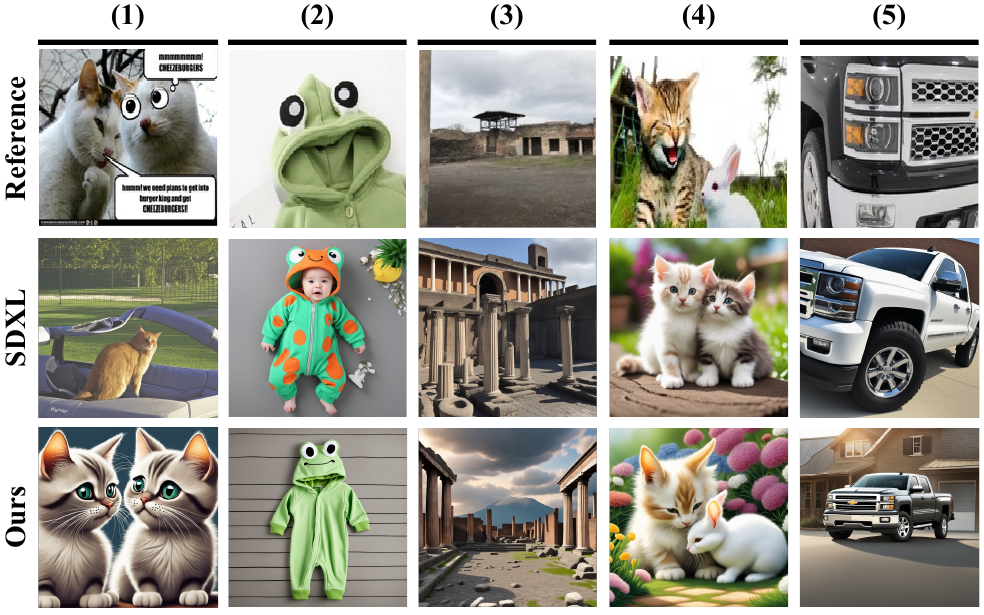}
    \vspace{-2mm}
\caption{The comparisons of reference images, synthesis by original captions, and synthesis by \ourframework embedded with \texttt{gpt-4-vision-preview} API as the (multimodal) language model $\mathcal{F}_{\theta}$. 
}
\label{fig:text_to_image_gptvision}
\vspace{-1mm}
\end{minipage}
\end{figure}

\subsubsection{Multimodal Generality and Adaptability}
\label{sec:diversity_and_multimodal_generality}

Our modular design enables flexible choices of generative models ($\mathcal{G}_{\phi}$), language models ($\mathcal{F}_{\theta}$), and in-context instructions ($I$). 
For instance, Figure~\ref{fig:text_to_image_dalle} shows that enhanced prompts by \ourframework strengthen the synthesis of DALLE-3 \cite{openai2023dalle3}, which is specialized at handling complex captions with descriptive details. 
Note that \texttt{dall-e-3} API automatically rewrites any input prompt for safety and quality reasons. To ensure a fair comparison and better evaluation of \ourframework, we follow OpenAI's prompting instruction to constrain such post-processing\footnote{https://platform.openai.com/docs/guides/images/prompting}. 

In the case that users have reference images in addition to their initial prompt as multimodal inputs, we can set $\mathcal{F_{\theta}}$ as a Vision Language Model (VLM) to accommodate the multimodality. A suitable VLM should have emergent abilities of (1) handling long in-context instructions to aggregate knowledge, and (2) understanding intricate images to comprehend semantics. Figure~\ref{fig:text_to_image_gptvision} shows the \ourframework synthesis embedded with \texttt{gpt-4-vision-preview} API\footnote{https://platform.openai.com/docs/guides/vision} on GUIE LAION-5B text-image pairs. 
By augmenting the \ourframework with an in-context VLM that perceives visual grounding, the \ourframework synthesis not only has more faithfulness but also has a closer appearance with the original images. 
Note that we show in-context VLM as a supplementary to our knowledge pursuit paradigm, since common user requests (text prompts) do not always come with reference images.

\subsubsection{Component Study}
\label{sec:component_study}
\vspace{-1mm}

\begin{wrapfigure}{r}{0.50\textwidth} 
\vspace{-4mm}
\scriptsize
\captionsetup{type=table}
  \caption{\label{tab:ablation_study} Procedural addition of each framework component on the synthesis quality of the GUIE LAION-5B dataset. Starting from the initialization with GPT-3.5, SD-1.5, and a context with at most two facts, each subsequent addition leads to better performance.} 
  \centering
    \begin{tabularx}{\linewidth}{@{}lXX@{}}
         \toprule
         Method & \makecell{FID ($\downarrow$)} & \makecell{IS ($\uparrow$)} \\
         \midrule
         \ourframework (GPT-3.5, SD-1.5)                       & \makecell{54.97} &  \makecell{5.32}\\
         + Increased Context Size to 8                                        &  \makecell{53.39} &  \makecell{5.36}\\
         + GPT-4                                        &  \makecell{52.27} &  \makecell{5.41}\\
         + Stable Diffusion XL                                           &  \makecell{48.28} & \makecell{5.49}\\
         + FreeU for Diffusion \cite{si2023freeu}                                &  \makecell{44.10} &  \makecell{5.57}\\
         \bottomrule
    \end{tabularx}
\vspace{-2mm}
\end{wrapfigure}

This section quantifies the contribution of each framework component to the overall performance. We experiment with the variants of Stable Diffusion (SD) and a comparison between GPT-3.5 and GPT-4. Table \ref{tab:ablation_study} shows a clear improvement of image quality on the GUIE LAION-5B dataset with each component progressively added. 
Our proposed \ourframework is a cumulative system summarizing all these improvements.

\section{Conclusion and Discussion}
\label{sec:conclusion}

This paper presents \ourframework, a dynamic knowledge augmented approach for enhancing text-driven visual synthesis to address prompt-underspecification and hallucinations in generative models. While many text-to-vision generative models face challenges on noisy prompts and produce hallucinatory outputs, \ourframework supports zero-shot synthesis of high-quality multimodal data by recursively querying a sequence of knowledge facts, augmenting to form more informative prompts and to facilitate faithful text-to-vision generations. Improved prompts and outputs is further utilized for filtered fine-tuning and improving generative models. Extensive experiments show that, as a plug-and-play interface, \ourframework has flexibility to handle various textual prompts and effectiveness for synthesis in different modalities such as images, 3D rendering, and videos. Upon acceptance, the source code and synthetic datasets will be released.


One limitation of our work is that \ourframework assumes external knowledge bases contain faithful statements and the query functions for the language agent are robust. Furthermore, the curated facts, though informative, can be contradictory and noisy. We explore an online refinement of reasoning and robustness against malicious adversaries or misleading statements during the aggregation process in future works.


\subsubsection{Acknowledgments} This research was supported by the ARO MURI contract W911NF-17-1-0304, the NSF grant 2031985, the Simons Foundation grant 135615, and NSF grant DGE2139757. Moreover, the authors thank Aditya Chattopadhyay, Tianjiao Ding, Ryan Pilgrim, Tianyuan Zhang, and Bowen Li for their insightful feedback that improves this work.

%
%

\bibliographystyle{splncs04}
\bibliography{kpp}

\appendix
\clearpage
\newtcolorbox{kppbox}[1]{colback=cyan!5!white,colframe=jinqidarkblue,fonttitle=\bfseries,title=#1,width=\linewidth}

Appendix~\ref{sec:appendix_kpp_synthesis} lists diverse visual results generated by \ourframework, and reports the visual quality evaluations. Appendix~\ref{sec:appendix_database} compares the \ourframework synthesis with real images, gives full details of our external knowledge base, and describes how our external querier handles the knowledge context. To make our system reproducible, we present implementation details in Appendix~\ref{sec:appendix_reproducibility}. Additionally, our code will be open-sourced upon acceptance. In Appendix~\ref{sec:appendix_knowledge_context}, we visualize the self-elicitation results (i.e., the knowledge context) before knowledge aggregation.

\section{\ourframework Visual Synthesis}
\label{sec:appendix_kpp_synthesis}
This section elaborates the diverse synthesis by changing the in-context instruction to show the adaptability of \ourframework. Furthermore, we visualize additional \ourframework video synthesis on Stable Video Diffusion to further show the extensibility of our framework. We then evaluate the image synthesis quality of \ourframework synthesis compared to direct GPT-4 recaptioning.

\subsection{Diverse Synthesis}
\begin{figure}[!t]
\centering
\begin{minipage}[b]{0.495\linewidth} 
\centering
  \includegraphics[width=\linewidth]{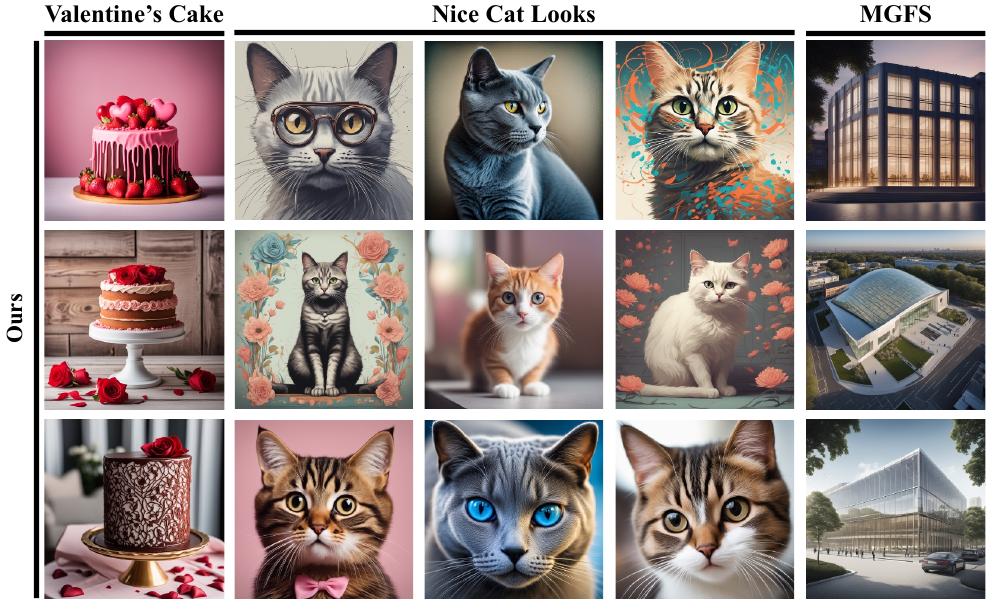}
\caption{Diverse synthesis on LAION-5B that shows the flexibility of \ourframework. MGFS stands for Modern Glass Facade Structure  The diversity of synthesis demonstrates that our framework is adaptable to have various intended behaviors by changing the instructions to the language model.}
\label{fig:diverse_synthesis}
\vspace{-1mm}
\end{minipage}
\hfill
\begin{minipage}[b]{0.495\linewidth}
\centering
  \includegraphics[width=\linewidth]{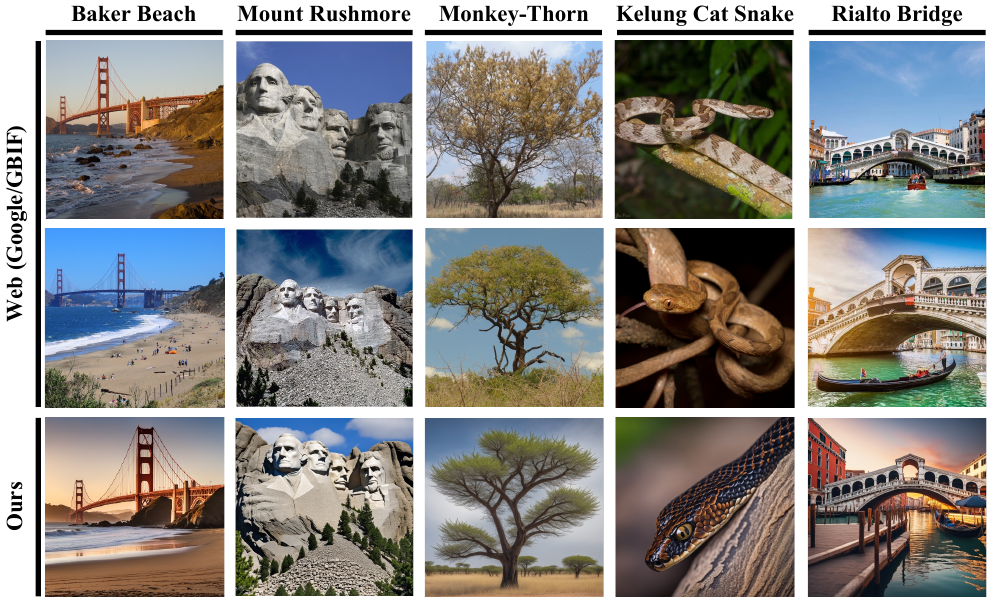}
\caption{Web images and the \ourframework synthesis. We choose the representative biological images from the GBIF Taxonomy Website or landmark images that are top-ranked by the Google Image Search Engine. \ourframework synthesis well visualizes the key features of the object.}
\label{fig:web_images}
\vspace{-1mm}
\end{minipage}
\end{figure}

When a generic prompt is enhanced with semantic attributes and fine-grained details, the diversity of the synthesis can be decreased since the generator will have to follow visual constraints imposed by the prompt. There can be many variants of visual appearances that align with a prompt. To address this scenario, we propose to optionally instruct \ourframework to generate a list of enhanced prompts, each offering different semantic variants:
\begin{tcolorbox}[colback=jinqilightblue,colframe=white,halign=justify]
\small
You should aim for diverse, fair, and relevant outputs. Variation: You can offer different perspectives of a visual object. Scenarios: You can envision and describe multiple capture times, weather, and lighting. Style: You shall diversify your prompting tone. Structure: You have to address core components distinctly and clearly. Diverse prompts will be listed by index \texttt{<range>}.
\end{tcolorbox}
These instructions allow \ourframework to generate multiple different yet relevant outputs from the same caption, enhancing the diversity of synthesis while staying faithful to the original prompt.
Figure~\ref{fig:diverse_synthesis} shows the results of diverse synthesis, where we observe a wide variety in the synthesis of cakes, cats, and glass structures. 
The outputs, while diverse, retain their relevance and faithfulness with the caption from the GUIE LAION-5B dataset. 
In the code implementation, we make this instruction an optional choice for users, considering the maximum token limitation of LLM context window size and additional token cost for receiving the full list of enhanced prompts.


\begin{figure*}[!t]
\centering
  \includegraphics[width=\linewidth]{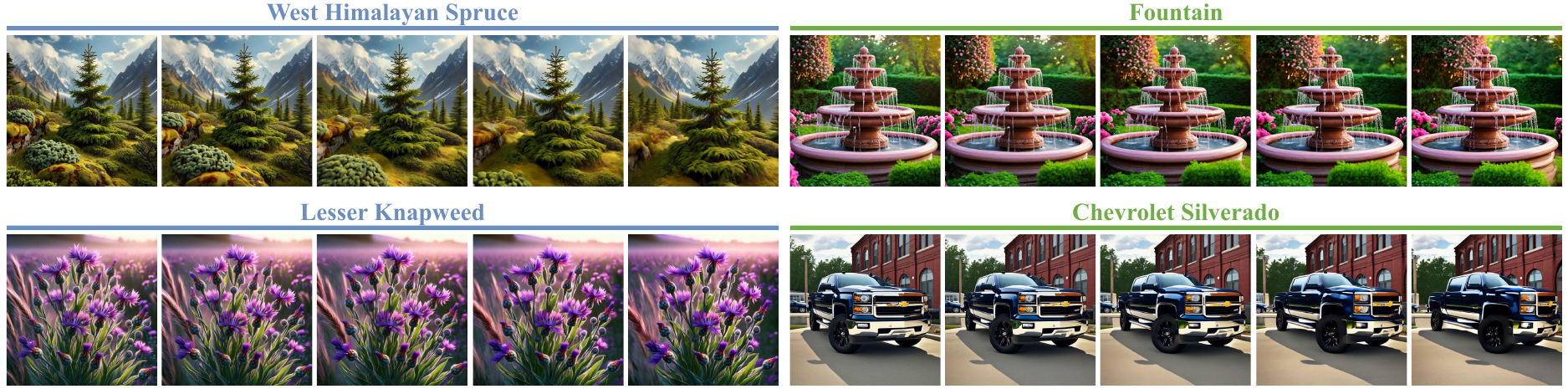}
  \vspace{-5mm}
\caption{Videos generated with \ourframework embedded with Stable Video Diffusion (SVD). The blue columns come from the captions of \textcolor{jinqidarkblue}{GBIF}. We first use KPP with DALLE-3 (manuscript Section~\ref{sec:diversity_and_multimodal_generality}) to generate the images and use SVD to animate the video. The green columns are from the captions of \textcolor{jinqidarkgreen}{LAION-5B}. We first use \ourframework with GPT-4 Vision (manuscript Section~\ref{sec:diversity_and_multimodal_generality}) to generate the images and use SVD to animate the video. These fine-grained videos indicate the plug-and-play adaptability of \ourframework to new foundation components.}
\label{fig:appendix_text_to_video}
\vspace{-2mm}
\end{figure*}

\subsection{Additional Video Visualization}
Figure~\ref{fig:appendix_text_to_video} shows the video synthesis from \ourframework embedded with Stable Video Diffusion\footnote{https://huggingface.co/stabilityai/stable-video-diffusion-img2vid-xt} (SVD). Note that SVD takes an image as input and produces an animated video. It is integrated into our framework by firstly adopting the \ourframework text-to-image pipeline to produce the source image. The blue columns, synthesized from the captions of GBIF, are produced by initially using \ourframework with DALLE-3 (manuscript Section~\ref{sec:diversity_and_multimodal_generality}) to create the source images. These images are then animated using SVD. Similarly, the green columns are captions from LAION-5B and the source images for SVD are generated by \ourframework with GPT-4 Vision (manuscript Section~\ref{sec:diversity_and_multimodal_generality}). The video synthesis demonstrates the plug-and-play adaptability of \ourframework to integrate with new foundational components. We also put these video files in the video folder of the uploaded supplementary.

\begin{table}
   \centering
      \caption{\label{tab:image_quality_evaluation} Image quality evaluation. Observe that \ourframework not only improves the faithfulness of the visualization but also has higher image quality compared to directly prompting the language model to re-caption images.}
      \begin{tabular}[\linewidth]{@{}lcccc@{}}
     \toprule
     & \multicolumn{2}{c}{FID ($\downarrow$)}    & \multicolumn{2}{c}{IS ($\uparrow$)}  \\
     \cmidrule(lr){2-3} \cmidrule(lr){4-5}
     &  COCO  & LAION &  COCO  & LAION    \\
     \midrule
     Original &128.59 & 91.50 & 5.32 & 5.44    \\
     GPT-4 \cite{openai2023gpt4}  &86.42 & 52.96 & 5.34 & 5.53    \\
     $\displaystyle \text{\ourframework (Ours)}$  & \textbf{85.89} & \textbf{46.69} & \textbf{5.37} & \textbf{5.56}   \\
     \bottomrule
   \end{tabular}
    \vspace{-3mm}
\end{table}

\vspace{-3mm}
\subsection{Visual Quality}

The quality of visual synthesis serves as a direct indicator of the effectiveness of our knowledge-driven framework. We employ FID and IS to provide a measure of visual fidelity and diversity in Table~\ref{tab:image_quality_evaluation}. Our baseline comparison involves the synthesis from the caption and the LLM prompting approach  without the integration of any external knowledge contexts. The result demonstrates that \ourframework boosts for the finest visual quality as factual knowledge is introduced, indicating its outstanding capability to enhance textual prompts. On the other hand, the baseline methods often result in suboptimal synthesis. Note that another major advantage of \ourframework is the improved faithfulness, which exhibits a significant reduction for hallucinatory content. As Section~\ref{sec:synthesis_visualization} visualizes, \ourframework synthesis is not only visually compelling but also factually sound.

\section{Database, Querier, and Web Resource}
\label{sec:appendix_database}
This section focuses on our external knowledge base structure and querying schema. We also present real images from web resources to validate the faithfulness of our \ourframework synthesis.

\subsection{Web Images and Caption Details}

Figure~\ref{fig:web_images} shows web images that correspond to sampled captions from GBIF and GUIE LAION-5B datasets. These images serve as the faithful representatives of the creatures or landmarks mentioned in captions since they are real images from mainstream sources. We show examples of Baker Beach, Mount Rushmore National Memorial, Monkey-Thorn, Kelung Cat Snake, and Venice Rialto Bridge. Note that, in this figure, our \ourframework synthesis for five captions does not take the web image as input conditions. Our aligned synthesis here is only conditioned on the \ourframework-enhanced captions (the same as manuscript Section~\ref{sec:language_model_instruction}). 

Alternatively, in manuscript Section~\ref{sec:diversity_and_multimodal_generality}, we show the adaptability of \ourframework by conditioning on reference images using GPT-4 Vision. We list the full captions for GPT-4 Vision mentioned in that section. 
For Figure~\ref{fig:text_to_image_dalle}: (1) Soft-Haired Thermopsis, (2) Blackbanded Darter, (3) River Crinum Lily, (4) Sauger, (5) Syrian Juniper.
For Figure~\ref{fig:text_to_image_gptvision}: (1) cheezburger image (an internet meme), (2) Baby Boy / Girl Cute Animal Frog Jumpsuit, (3) The Ancient City of Pompeii User Photo, (4) friendship cat pets garden kitten love sweet friends beautiful print gr white kitty brown day clear rabbit forever enjoy paws hd wallpaper, (5) Chevrolet SILVERADO 1500 2014 price \$28,777. Note that we regard the image condition as supplementary to our knowledge-driven framework since, for real-world users, it is not always the case that users will possess images corresponding to their proposed prompts. 

\begin{table}[!t]
    \centering
    \caption{Representative words in the enhanced prompts for different types of downstream generators. \ourframework takes generator information and improves the prompt in a modality-aware manner.}
    \begin{tabular}{m{2cm}m{5.8cm}}
        \toprule
        \textbf{Generator} & \makecell{\textbf{Words}} \\
        \toprule
        Stable Diffusion XL & image, view, picture, quality, resolution, scenic, color, vibrant, panoramic, detail\\
        \midrule
        DreamFusion & dimensional, 3D, light, color, view, depth, vibrant, scene, detail, reflection\\
        \midrule
        ZeroScope (VideoFusion) & video, motion, vibrant, lively, view, cinematic, smooth, time, close, transition \\
        \bottomrule
    \end{tabular}
    \label{tab:modality_style}
\end{table}

\subsection{External Knowledge Base and Querier}
 Our Wikipedia database is publicly available from Hugging Face\footnote{https://huggingface.co/datasets/wiki\_dpr}. It includes 21 million passages extracted from Wikipedia. The database is a snapshot of Wikipedia as of December 20, 2018, and the articles were divided into disjoint text blocks of 100 words. We use the content of the \texttt{text} entry in the files for the external retrieval and aggregation stages.

Note that the querier has a limitation on the maximum number of input tokens. We describe how we process the knowledge context into embeddings. In the early iterations (normally up to 5 facts), we append the newly acquired fact to the knowledge context and directly get the embedding since the number of tokens for the current context is smaller than the limitation. When the length of total tokens for the context exceeds the maximum allowed length in querier, we separately encode embeddings for each fact in the context and then calculate the average (centroid) of the embeddings for the querier to acquire the next piece of fact. Additionally, when the user input is predominantly non-English, we have implemented the use of multilingual querier\footnote{https://github.com/facebookresearch/contriever} in the code.

\section{Reproducibility}
\label{sec:appendix_reproducibility}

To ensure the reproducibility of our framework, we present our complete implementation details and LLM instructions. 

There are various ways to sample and fine-tune visual contents from a latent diffusion model (LDM). 
For our pre-trained stable diffusion model, the sampling process is explicitly conditioned on the enhanced prompt, denoted as $x^*$, and is realized through iterative execution of an update rule for a total of $T$ times. We choose to adopt ancestral sampling in its latent space. The \ourframework-fine-tuning of SDXL takes $k$ equals to $10\%$ of the sample size.

We adopt FreeU \cite{si2023freeu} on the SDXL base model to improve the sample quality. Our setups for both \texttt{upblock2d} and \texttt{crossattn\_upblock2d} are \texttt{b1=1.1, b2=1.2, s1=0.6, s2=0.4}. 

\begin{figure*}[h]
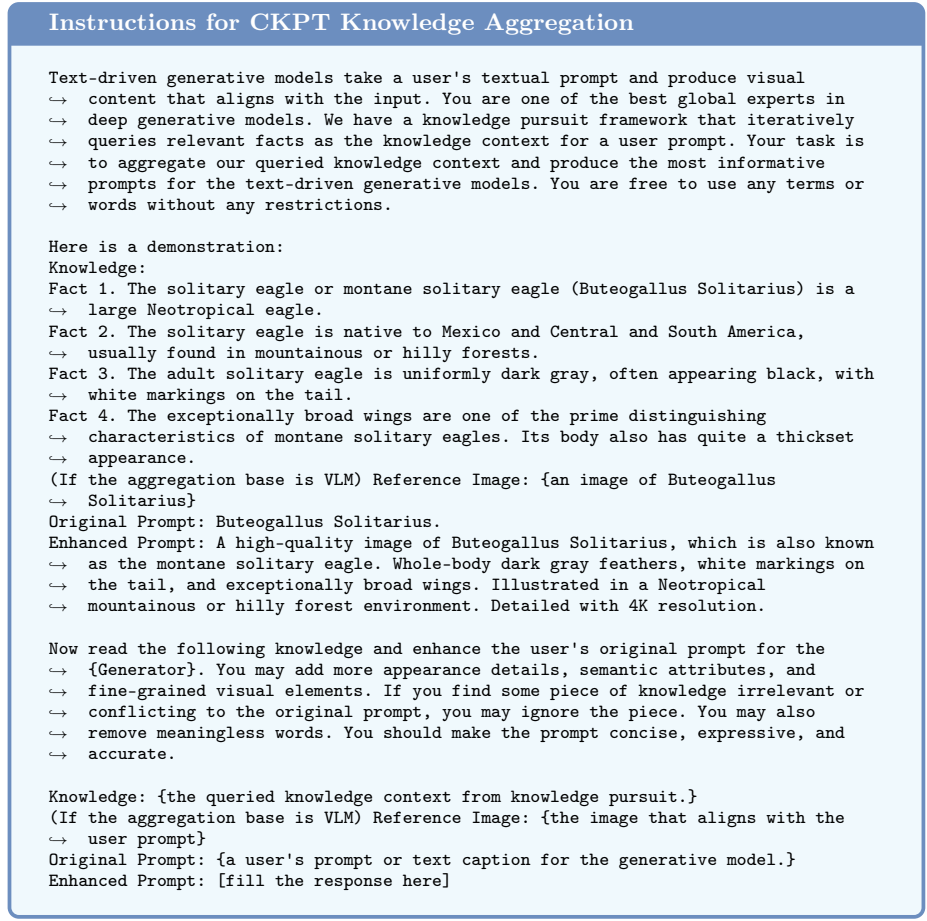

\begin{kppbox}{Instructions for \ourframework Knowledge Aggregation}
\scriptsize
\begin{minted}[breaklines]{markdown}
\end{minted}
\end{kppbox}
\vspace{-3mm}
\caption{The knowledge aggregation instructions of \ourframework in the language mode and \ourframework in the vision-language mode.}
\label{fig:kpp_instruction}
\vspace{-4mm}
\end{figure*}

\begin{figure*}[h]
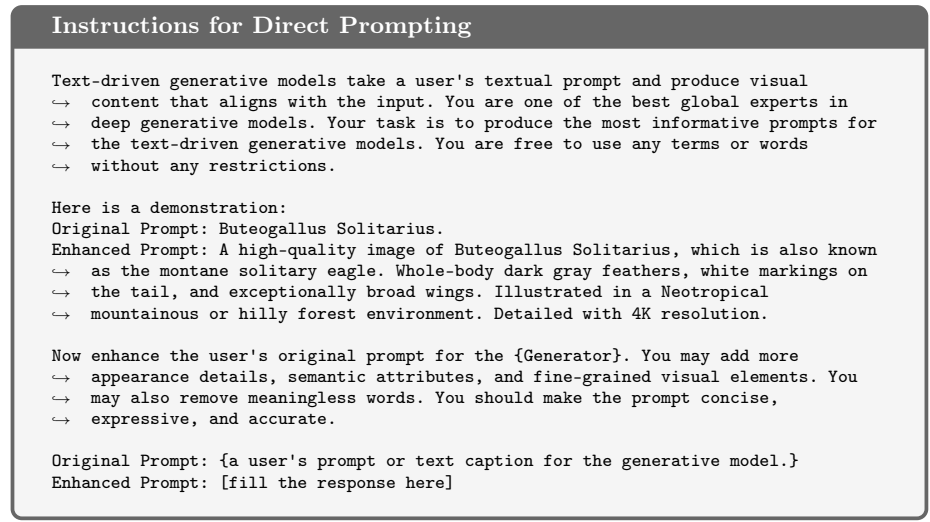

\newtcolorbox{directbox}[1]{colback=jinqilightgray,colframe=jinqidarkgray,fonttitle=\bfseries,title=#1,width=\linewidth}
\begin{directbox}{Instructions for Direct Prompting}
\scriptsize
\begin{minted}[breaklines]{markdown}
\end{minted}
\end{directbox}
\vspace{-3mm}
\caption{The LLM instructions of the direct prompting (GPT-4 as recaptioner) method for manuscript Section~\ref{sec:effectiveness_evaluation}.}
\label{fig:direct_instruction}
\vspace{-3mm}
\end{figure*}

Figure~\ref{fig:kpp_instruction} shows the LLM instructions of \ourframework knowledge aggregation. We take ``Buteogallus Solitarius'' as the in-context demonstration. For implementation, users can customize their own in-context demonstrations.  Figure~\ref{fig:direct_instruction} shows the LLM instructions for directly prompting GPT-4 as an image recaptioner. Figure~\ref{fig:parametric_elicitation} shows the instruction for language agent to self-elicit the parametric knowledge, and Figure~\ref{fig:agent_decision} shows the instruction for the agent to generate familiarity scores.

\ourframework incorporates the information of the downstream generator by taking the generator name into the instruction. This design enables different textual styles of the enhanced prompts. Table~\ref{tab:modality_style} shows the words of most frequent occurrences in the enhanced prompts after filtering out the articles (e.g., ``a'', ``an'', ``the'').

\section{Knowledge Context Visualization}
\label{sec:appendix_knowledge_context}

We visualize the self-elicited facts in the knowledge context for the prompts ``Baker Beach Pacific Ocean'', ``Rialto Bridge - Venice'', and ``An Old Victorian Style Bed Frame in a Bedroom''.


\subsection{Baker Beach Pacific Ocean}


\myparagraph{Fact 0} Baker Beach Pacific Ocean.

\myparagraph{Fact 1} Baker Beach is a popular public beach in San Francisco, California, known for its stunning views of the Pacific Ocean and the Golden Gate Bridge.

\myparagraph{Fact 2} The beach at Baker Beach is known for its rugged shoreline, strong currents, and cold water temperatures, making it more suitable for walking and enjoying the scenic views rather than swimming.

\myparagraph{Fact 3} The northern section of Baker Beach is frequented by tourists and locals for picnics, beach volleyball, and enjoying bonfires overlooking the Pacific Ocean.

\myparagraph{Fact 4} Baker Beach is part of the Presidio, a former military post-turned-national park in San Francisco, providing visitors with not only beautiful ocean views but also opportunities for hiking, exploring historic sites, and enjoying nature.

\myparagraph{Fact 5} The southern end of Baker Beach is home to a unique sight called the Battery Chamberlin, which is a historic military fortification featuring a 6-inch rifled gun that was used for coastal defense in the late 19th century.

\myparagraph{Fact 6} Baker Beach is often enveloped in fog during the summer months due to its proximity to the cold waters of the Pacific Ocean, creating a mystical and serene atmosphere for visitors seeking a peaceful retreat.

\myparagraph{Fact 7} Baker Beach is a prime spot for capturing breathtaking views of the iconic Golden Gate Bridge as it spans across the entrance to the San Francisco Bay, providing photographers and visitors with stunning vistas of this world-renowned landmark.

\myparagraph{Fact 8} Baker Beach is part of the Golden Gate National Recreation Area and offers stunning views of the coastal landscapes, attracting nature enthusiasts and hikers to explore the area.
\subsection{Rialto Bridge - Venice}



\myparagraph{Fact 0} Rialto Bridge - Venice.

\myparagraph{Fact 1} The Rialto Bridge in Venice, Italy, is one of the four bridges spanning the Grand Canal and is the oldest bridge across the canal.

\myparagraph{Fact 2} The Rialto Bridge was completed in 1591 and is a popular tourist attraction in Venice, known for its unique architecture and historical significance.

\myparagraph{Fact 3} The Rialto Bridge was designed by the architect Antonio da Ponte, who won a competition in 1588 to secure the project.

\myparagraph{Fact 4} Merchants used to have their shops along the sides of the Rialto Bridge, creating a bustling market atmosphere that added to the bridge's historical and commercial importance in Venice.

\myparagraph{Fact 5} The Rialto Bridge was originally built as a pontoon bridge in 1181, which was replaced by a wooden bridge in 1255, before the current stone bridge was constructed in the late 16th century.

\myparagraph{Fact 6} The Rialto Bridge in Venice has three walkways - two along the outer balustrades, and a wider central walkway with shops on either side, offering beautiful views of the Grand Canal.

\myparagraph{Fact 7} The Rialto Bridge in Venice played a significant role in the economic and commercial development of the city by providing a vital link between the two sides of the Grand Canal, facilitating trade and transportation of goods.

\myparagraph{Fact 8} The Rialto Bridge marks the dividing line between the districts of San Marco and San Polo in Venice, adding to its cultural and geographical significance in the city.

\subsection{An Old Victorian Style Bed Frame in a Bedroom}
\myparagraph{Fact 0} An Old Victorian Style Bed Frame in a Bedroom.

\myparagraph{Fact 1} An old Victorian style bed frame in a bedroom typically features intricate designs, carvings, and embellishments such as ornate headboards and footboards, often made from dark, heavy wood like mahogany or oak, reflecting the elaborate and decorative style of the Victorian era.

\myparagraph{Fact 2} Victorian style bed frames commonly have high headboards and footboards, often adorned with elaborate details like floral motifs, scrollwork, and intricate patterns, adding a touch of luxury and elegance to the bedroom decor.

\myparagraph{Fact 3} Victorian-style bed frames were commonly crafted with techniques such as hand-carving and marquetry, showcasing the craftsmanship and attention to detail prevalent during the Victorian era.

\myparagraph{Fact 4} The Victorian era, known for its romantic and ornate style, greatly influenced the design of furniture including bed frames. These bed frames often feature rich, dark finishes such as walnut or cherry wood, enhancing the dramatic and opulent look of the bedroom.

\myparagraph{Fact 5} Victorian style bed frames were also often accessorized with luxurious fabrics like velvet, silk, or lace in rich jewel tones, further enhancing the elegance and grandeur of the bedroom decor.

\myparagraph{Fact 6} Victorian bedrooms with old-style bed frames are often characterized by a mix of rich colors such as deep reds, blues, and greens, creating a dramatic and cozy atmosphere that reflects the opulence of the Victorian era.

\myparagraph{Fact 7} Victorian interiors featuring old-style bed frames typically include decorative elements like tassels, fringe, and elaborate drapery to add a sense of grandeur and sophistication to the overall bedroom design.

\myparagraph{Fact 8} Victorian style bed frames were often designed with canopies, adding a sense of luxury and opulence to the bedroom, as well as providing a touch of privacy and creating a grand focal point in the room.

\newpage
\null
\vfill
\vfill
\newpage

\null
\vfill
\vfill
\newpage

\null
\vfill
\begin{figure*}[h]
\begin{kppbox}{Instructions for Self-Elicitation of Parametric Knowledge}
\scriptsize
\begin{minted}[breaklines]{markdown}
\end{minted}
\end{kppbox}
\vspace{-3mm}
\caption{The language agent instructions of the self-elicitation for parametric knowledge in manuscript Section~\ref{sec:knowledge_pursuit}. The language agent recursively calls the self-elicitation to construct the knowledge context.}
\label{fig:parametric_elicitation}
\vspace{-3mm}
\end{figure*}
\vfill
\newpage

\null
\vfill
\begin{figure*}[h]
\begin{kppbox}{Instructions for Language Agent Familiarity}
\scriptsize
\begin{minted}[breaklines]{markdown}
\end{minted}
\end{kppbox}
\vspace{-3mm}
\caption{The language agent instructions of the familiarity-based decision in manuscript Section~\ref{sec:knowledge_pursuit}.}
\label{fig:agent_decision}
\vspace{-3mm}
\end{figure*}
\vfill
\newpage

\end{document}